\documentclass[letterpaper]{article} 
\usepackage{aaai2026}  
\usepackage{times}  
\usepackage{helvet}  
\usepackage{courier}  
\usepackage[hyphens]{url}  
\usepackage{graphicx} 
\usepackage{amsmath, amssymb, amsthm}
\usepackage{natbib}  
\usepackage{caption} 
\usepackage{algorithm}
\usepackage{algorithmic}

\usepackage{newfloat}
\usepackage{listings}
\DeclareCaptionStyle{ruled}{labelfont=normalfont,labelsep=colon,strut=off} 
\floatstyle{ruled}
\newfloat{listing}{tb}{lst}{}
\floatname{listing}{Listing}
\newtheorem{prop}{Proposition} 
\title{Optimizing Urban Service Allocation with Time-Constrained Restless Bandits}
\author{
    Yi Mao,
    Andrew Perrault
}
\affiliations{
    The Ohio State University\\
    \{mao.496, perrault.17\}@osu.edu
}

\begin{document}

\maketitle

\begin{abstract}
Municipal inspections are an important part of maintaining the quality of goods and services. In this paper, we approach the problem of intelligently scheduling service inspections to maximize their impact, using the case of food establishment inspections in Chicago as a case study. The Chicago Department of Public Health (CDPH) inspects thousands of establishments each year, with a substantial fail rate (over 3,000 failed inspection reports in 2023).  
To balance the objectives of ensuring adherence to guidelines, minimizing disruption to establishments, and minimizing inspection costs, CDPH assigns each establishment an inspection window every year and guarantees that they will be inspected exactly once during that window. Meanwhile, CDPH also promises surprise public health inspections for unexpected food safety emergencies or complaints.
These constraints create a challenge for a restless multi-armed bandit (RMAB) approach, for which there are no existing methods.
We develop an extension to Whittle index-based systems for RMABs that can guarantee action window constraints and frequencies, and furthermore can be leveraged to optimize action window assignments themselves. Briefly, we combine MDP reformulation and integer programming-based lookahead to maximize the impact of inspections subject to constraints. A neural network-based supervised learning model is developed to model state transitions of real Chicago establishments using public CDPH inspection records, which demonstrates 10\% AUC improvements compared with directly predicting establishments' failures. Our experiments not only show up to 24\% (in simulation) or 33\% (on real data) objective improvements resulting from our approach and robustness to surprise inspections, but also give insight into the impact of scheduling constraints.
\end{abstract}


\section{Introduction}
Cities perform inspections in order maintain the quality of goods and services available to their residents. In this work, we consider the inspection of food establishments. The city's goal can be summarized as: keep as many establishments as possible in the inspection-passing state while guaranteeing that each establishment is inspected at a certain frequency (e.g., exactly once per year or between once and twice per year). The former objective can be viewed as supporting the average citizen by maximizing the probability that a random establishment is in an inspection-passing state. The second provides a worst-case guarantee: there is a limit on how long ago any establishment would have been inspected and a limit on how much disruption establishments experience. The city must achieve these goals subject to a budget on the number of inspections that can be performed per unit time.

Restless multi-armed bandits (RMABs)~\cite{whittle_1988} are \emph{almost} a natural fit for this problem. They describe a sequential decision problem where an agent (the city) acts on a large population of independently evolving Markov decision processes (the establishments), which describe each establishment's propensity to stay in the inspection-passing state. While RMABs are highly suitable for maximizing the inspection-passing probability for each establishment, existing approaches (e.g., ~\cite{deadline,fair_bandit,li2023avoiding}) fail to support the constraint structure required for establishment scheduling, where each arm's pulls must satisfy an \emph{ex-post} frequency constraint.

In this paper, we develop new methods that allow us to solve these problems by combining Whittle index theory and integer programming. Despite their use of integer optimization, which is often slow, our methods are highly scalable, often due to the unimodularity of the arm scheduling optimization. One key new question that arises is how to assign an inspection \emph{window} to each establishment. To minimize disruption, establishments are informed in advance of a set of contiguous time periods during which their inspection will occur. We show that, through an extension of our methods for RMABs under constraints, we can optimize the assignment of windows to arms and ensure that establishments cannot predict when their inspection will occur in the assigned window.

In experiments using synthetic data and real data from the Chicago Department of Public Health (CDPH), we evaluate the impact of both RMAB inspection scheduling and window optimization and find that a substantially higher reward can be achieved through optimization.

Our contributions are as follows:
\begin{itemize}
\item \textbf{RMABs under service constraints.} We introduce a method for optimizing inspections under given scheduling window and frequency constraints. To upper bound the number of inspections during a time window, we integrate the service windows into the structure of the MDP by introducing new timing states. This approach is structural and ensures that window constraints are never violated.
Frequency constraints (a certain number of inspections should occur over a longer period of time) present challenges because the Whittle index heuristic is greedy and does not look into the future. We introduce an integer programming-based planner that aims to maximize the sum of future Whittle indices subject to constraints. We find this integer program to be highly scalable and show that the constraint matrix is totally unimodular in some cases.
\item \textbf{Optimizing inspection windows.} We show that our method for optimizing inspections under fixed constraint structures can be leveraged to optimize the timing of inspection windows themselves. This optimization is done carefully to avoid leaking excess information about the time of specific inspections to the establishments based on their window assignments. The optimization of inspection windows turns out to be critical to produce higher rewards in our experiments, as it provides more flexibility to the scheduler.
\item \textbf{Empirical evaluation and impact/robustness assessment.}
For the CDPH, we develop a machine-learning approach to estimate the parameters of each establishment based on historical data. We evaluate reward and computation time in both real and synthetic data. In our tested cases, we find that both explicitly modeled constraints and optimized action windows bring substantial advantages compared with ad hoc methods for handling constraints. Considering some urban emergent service would break the original schedulings, we evaluate our method's robustness by adding unexpected mandatory actions and find our method is more robust than ad hoc plannings.
\item \textbf{Machine-learning model predicting transitions of MDP. }
There is one previous work predicting the food inspection failures directly using machine-learning models~\cite{foodforecast}. For the purpose of building MDPs for our RMAB, we instead train a neural network to predict the state transition probabilities. We found the model not only aligns with the requirements of RMAB, but also has the ability to predicting the inspection failures. It improves the AUC number by around 10\% compared with the XGBoost model proposed by~\citet{foodforecast}.
\end{itemize}
Collectively, our analysis advances the state of the art in RMAB planning under constraints as well as providing insight into the impact of optimizing city service schedules with improvements of up to 33\%.

\section{Related Work}
\paragraph{Food safety inspections.} 
In 2015, CDPH leveraged historical food inspection data and trained a supervised learning model to predict the probability that an inspection would uncover a critical violation~\cite{foodforecast}. ~\citet{kannan2019hindsight} independently analyzed the impact of prediction-driven scheduling. However, such models only consider one-shot predictions for critical violations and do not include the sequential aspect of scheduling. Fairness is of substantial interest in the provision of municipal services ~\cite{singh2022fair}. We consider fairness outside the scope of this paper, but a potentially interesting direction for future work.

\emph{Restless multi-armed bandits (RMABs).}
RMABs are PSPACE-hard in the worst case, but~\citet{whittle_1988} showed that a subclass of them, so-called indexable RMABs, admit an efficient asymptotically optimal solution. Particularly relevant classes of indexable RMABs are those that extend the machine maintenance problem families~\cite{some_family}, and scheduling problems for sensors~\cite{aoi}, wireless transmission~\cite{wireless_trans}, and health interventions~\cite{mate2020collapsing}. These RMABs are structured so the state of each process declines if it is not acted on, and differ in the details of action effect and what information is observed with or without an action. This work aims to develop techniques to integrate action constraints into RMABs of these types.


\emph{RMABs with Constraints.}
Several RMAB models have included constraints. In a project applying RMABs to assist maternal and child health via phone calls, a ``sleeping period'' for arms was enforced after they were pulled by the Whittle index heuristic~\cite{field_study} (See~\ref{sec:window}). It appears to have been enforced in an ad hoc manner, by blocking pulls that would have violated the constraint.
~\citet{deadline} deployed RMABs in a deadline scheduling setting and integrated the deadline constraints by adding dummy arms. Fairness is another setting where constraints can arise. ~\citet{fair_bandit} introduced the ProbFair policy, ensuring a strictly positive lower bound on the probability of being pulled at each time step while still satisfying the budget constraints. ~\citet{li2023avoiding}'s SoftFair balances the trade-off between the goals of having resources uniformly distributed and
maximizing cumulative rewards. They guarantee a long-term probability of each arm being pulled whereas ours ensures pulling frequencies for each arm strictly in a period. To the best of our knowledge, we are the first to consider action window and frequency constraints strictly for recurring tasks in a period.

\section{Preliminaries}

An RMAB consists of $N$ binary action MDPs (\emph{arms}). 
We define the $i$th two-action MDP~\cite{10.5555/528623} as a tuple $(\mathcal{S}_i, \mathcal{A}, P_i, R_i, s^{(0)}_i, \gamma)$. The discount factor $\gamma$ and action space $\mathcal{A}=\{0, 1\}$ are fixed across all MDPs. When the action 1 (resp. 0) is taken on an arm at time $t$, we refer to that arm as \emph{active} (resp. \emph{passive}). The rest are arm specific: $\mathcal{S}_i$ is the state space, $P_i: \mathcal{S}_i \times \mathcal{A} \rightarrow \Delta \mathcal{S}_i$ is the transition function, $s^{(0)}_i$ is the start state, and $R_i: \mathcal{S}_i \times \mathcal{A} \rightarrow \mathbb{R}$ is the reward function. Because there are only two actions, the transition function $P_i$ can be decomposed into an an active transition $P^{(1)}_i: \mathcal{S}_i \rightarrow \Delta \mathcal{S}_i$ and a passive transition $P^{(0)}_i: \mathcal{S}_i \rightarrow \Delta \mathcal{S}_i$.

A RMAB consists of $N$ binary action MDPs and a per-timestep budget constraint $k$. At each round $t$, the agent has a budget $k$, where $k \ll N$, meaning that at most $k$ arms can be ``pulled'', i.e., have their action set to 1. The MDP which is pulled transits actively and otherwise transits passively. Upon transitions, the rewards from all MDPs are collected and accumulated over time. The goal is to find an optimal policy $\pi^\star$ to maximize our total rewards---formally,
\begin{align}
     \pi^* = \arg\max_{\pi} J = \nonumber\\\arg\max_{\pi: \sum_i{\pi_i(S_{t}) \leq k} }\sum_i\sum_t \gamma^t R_i(s_{i,t}, \pi(S_{t})),
\end{align}
where $\pi_i(S_t) \in \mathcal{A}$ is the action selected by $\pi$ for arm $i$, $S_t \in \mathcal{S}_1 \times \ldots \times \mathcal{S}_N$ is the joint state of all arms at time $t$, and $s_{i, t} \in \mathcal{S}_i$ is the state of arm $i$ at time $t$.

\subsection{Whittle Indices}
General RMABs have an exponentially large state space and a combinatorially large action space. 
The Whittle index method provides tractability for some classes of RMABs~\cite{whittle_1988}. It works by computing a ``benefit of acting'' for each arm, called the \emph{Whittle index}. The \emph{Whittle index heuristic} then acts on the $k$ arms with highest Whittle indices.

To calculate the Whittle index for each arm, we search over ``subsidies'' for the passive action \emph{m}. Formally, the subsidy $m$ modifies the reward function $R_i$ into $R^{(m)}_i$:
\begin{align}
    R^{(m)}_i(s_{i}, 0) = R_i(s_i) + m; R^{(m)}_i(s_{i}, 1) = R_i(s_i).
\end{align}
The goal is to identify the smallest subsidy $m$ such that, for the current state $s_{i,t}$, the long-term rewards for the passive and active actions are the same. To formalize this, we first define the $Q$ function for arm $i$ under subsidy $m$:
\begin{align}
Q_i^{(m)}(s_i, a) &= R_i^{(m)}(s_i, a) + \nonumber\\ &\gamma \max_{a' \in \mathcal{A}} \sum_{s'_i \in \mathcal{S_i}} P_i(s_i' | s_i, a) Q_i^{(m)}(s'_i, a').
\end{align}

\paragraph{Definition} The Whittle index for state $s_{i,t}$ is the smallest \emph{m} which makes it equally optimal to take the active and passive actions: 
\begin{small}
\begin{align} 
    w(s_{i,t}) = \inf_m \left\{ m: Q_i^{(m)}(s_{i,t}, a = 0) \geq Q_i^{(m)}(s_{i,t}, a = 1) \right\}.
\end{align}
\end{small}

For the Whittle index heuristic to have asymptotic optimality guarantees, each arm must satisfy a technical condition called \emph{indexability}~\cite{whittle_1988}. Intuitively, indexability says that, as $m$ increases, the optimal action can only switch to passive and cannot switch back to active. Let $W_i^{(m)}$ be the set of states for which $Q_i^{(m)}(s_{i,t}, a = 0)\geq Q_i^{(m)}(s_{i,t}, a = 1)$, i.e., the passive action has an equal or higher return than the active action.

\paragraph{Definition} (Indexability). An arm is said to be indexable if $W_i^{(m)}$ is non-decreasing in \emph{m}, i.e., for any $m_1, m_2 \in \mathbb{R} $ such that $m_1 \leq m_2 $, we have $W_i^{(m_1)} \subseteq W_i^{(m_2)}$. An RMAB is indexable if every arm is indexable.

\section{Satisfying Inspection Constraints}
\label{sec:satisfying}

We study two types of action constraints that arise in the motivating food establishment inspection problem. 
We begin by defining a sample RMAB with domain-motivated constraints (Sec.~\ref{sec:exampleRMAB}).
Window constraints specify an action window where the arm is allowed be acted on (Sec.~\ref{sec:window}). Frequency constraints specify a minimum number of actions each arm must receive over a period of time (Sec.~\ref{sec:frequency}).

\subsection{Motivating Inspection RMAB}\label{sec:exampleRMAB}
Motivated by the food establishment setting, we define a model RMAB to which we will add action constraints. This RMAB can be viewed as a collapsing bandit~\cite{mate2020collapsing} or a resetting bandit~\cite{resource}, and both have indexability guarantees. Each establishment has an unobserved binary state that is either 1 (i.e., inspection passing) or 0 (i.e., inspection failing). When we act on the establishment, we assume that it is restored to the passing state and define the reward function to be 1 for each time period the establishment is in the passing state and 0 otherwise. We consider time periods as months---each establishment needs to be inspected once a year and will have a two-month period where this inspection can occur.

As the true states are not directly observable, each arm is a partially observed Markov decision process (POMDP)~\cite{pomdp}. We can rewrite the POMDP as a fully observed belief-state MDP, allowing for direct representation as an RMAB.

For the underlying MDP, we assume passive transitions $P_i^{(0)}$ and active transitions $P_i^{(1)}$ as follows:
\begin{align*}
    P_i^{(1)} = 
    \left( 
        \begin{array}{cc}
           1  & 0 \\
           1   & 0
        \end{array}
    \right), \hspace{10pt}
    P_i^{(0)} = 
    \left( 
        \begin{array}{cc}
           P_i^{(00)}  & P_i^{(01)} \\
           P_i^{(10)}   & P_i^{(11)} 
        \end{array}
    \right)
\end{align*}
Each establishment has its own passive transition probabilities and all share the same action impacts---actions always restore the establishment to the passing state in the next timestep.

Converting this POMDP to a belief-state MDP yields a set of belief states that are reachable from the passing state $b_1=[0, 1]$ (as a column vector), i.e., $(P_i^{(0)})^t b_1$, where $t$ is any non-negative integer. In practice, the number of states needed to model belief dynamics precisely enough is dependent on the rate of MDP mixing. A faster mixing MDP will reach its stationary state faster and require fewer states---once we are sufficiently close to the stationary state, we can have the state transition to itself. The resulting belief-state MDP has a chain structure as shown in Fig~\ref{fig:mdp} and resets to the head of the chain when the active action is taken. 

Collapsing bandits generalize this setting by allowing $P_i^{(1)}$ to vary per arm, resulting in a two-chain structure. In general, our methods will also apply to this setting with minor modifications.

\subsection{Action Windows and MDP Encoding}\label{sec:window}
We use action windows as an exemplar for the family constraints where the constraint can be directly encoded into the RMAB structure, i.e., a vanilla RMAB with an action window constraint can be rewritten as a vanilla unconstrained RMAB a different arm structure. This is in some sense the ideal way to add constraints---we can apply whatever existing state-of-the-art algorithm directly.

To add action windows to the MDP structure, we add two pieces of information to the states (in addition to the belief state $b \in [0,1]$), and modify the transitions to remove the impact of actions outside the window.
\begin{itemize}
    \item $t$: the current timestep. In our motivating example, we can use $t \textrm{ mod } 12$, as the inspection window for each establishment is at the same time each year. Alternatively, if the windows are not periodic, this can be replaced with a pair of counters, with one indicating the remaining time in the current window and the other referring the time until the next window.
    \item $m$: A counter for the number of actions remaining in the action window. When the process enters the action window, this is set to the total number of active actions allowed during the window (in our motivating example, this is 1). Each active action decrements the counter by 1. If the counter is zero, the active action is still available, but it has the same transitions as the passive action.
\end{itemize}

\begin{figure}[t]
\centering
\includegraphics[width=1\columnwidth]{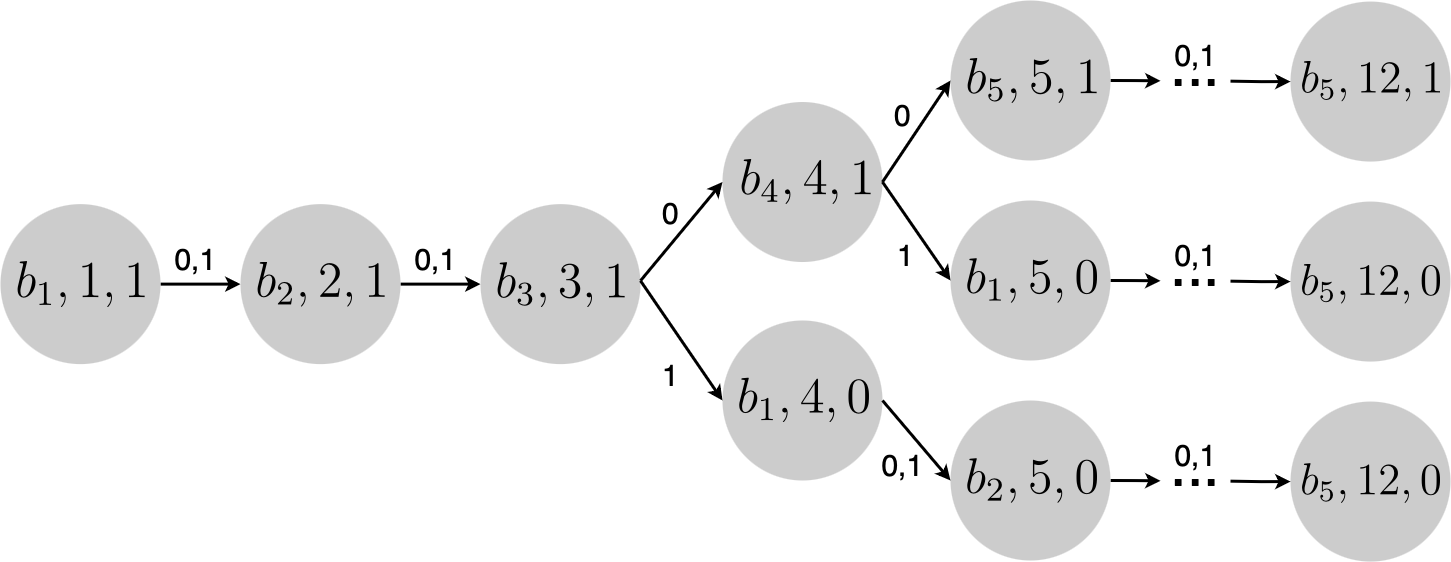} 
\caption{A example portion of the MDP after encoding the action window constraint. Suppose we have 5 belief states $(b_1, ..., b_5)$, an action window at months 3 and 4, and 12 months between action windows. 0 is the passive action, 1 is the active action. After $(b_5, 12, 0)$ is reached, a new chain begins at $(b_5, 1, 0)$ (not shown).}
\label{fig:mdp}
\end{figure}

As shown in Figure~\ref{fig:mdp}, such an encoding increases the number of states in the MDP. The number of states is increased by a factor of $O(LM)$, where $L$ is the number of counter values required to track when the window is active, and $M$ is the total number of actions allowed during the window. For the motivating RMAB, this increase is by a factor of 14. But such encoding is also applicable to more complex situations: An arm has multiple action windows and multiple inspections allowed per window. By design, the Whittle heuristic ensures no arms outside the action window are selected and no ill effect even if selected by accident. We provide a precise description of the new MDP in the appendix.  All new transitions are deterministic (probability = 1).



Adding action windows to the MDP encoding will cause the Whittle index to increase when the end of a window is reached. This makes it more likely that an arm will be pulled before its window expires. Nevertheless, we begin to encounter the limits of the greedy Whittle index heuristic. For example, if we have several arms that have action windows that end on the same timestep, we may miss some action opportunities without planning ahead. In the next section, we develop a method for planning with lookahead, which will allow us to enforce frequency constraints as well as optimize timing of actions subject to window constraints.

\paragraph{Indexability.} We are not aware of an existing class of indexable RMABs that includes the action window MDPs with counters that we define in this section. We empirically check for indexability through tracking the set of passive states as the subsidy changes and find no violations.

\subsection{Frequency Constraints and Lookahead}\label{sec:frequency}

It is possible to enforce maximum action limits via editing the individual MDPs, but it is not possible to enforce minimums this way. In the motivating RMAB, we want to enforce the constraint that each establishment is inspected exactly once or multiple times per year since in the food inspection task, the authority has responsibilities to inspect every food establishment and never skip one. To enforce this kind of frequency constraint, we will replace the Whittle index heuristic with a sequential planning component that aims to maximize the sum of indices of pulled arms over a lookahead window, not just in the next timestep.

We begin with the case where each arm needs to be pulled exactly one time over the lookahead window (and later relax this). In the motivating RMAB, this window will be one year. Formally, we let $a_{i,t}$ be whether arm $i$ is pulled at time $t$ and $w_{i,t}$ be the Whittle index of arm $i$ at time $t$. 
These Whittle indices can come from an RMAB with any encoded constraints, such as those in the previous section.
We seek to maximize $\sum\limits_{i=1}^{N}\sum\limits_{t=1}^{T} a_{i,t}w_{i,t}$, subject to the following constraints:
\begin{enumerate}
    \item $\sum\limits_{i=1}^{N}a_{i,t} \leq k$: only $k$ arms can be pulled in each timestep.
    \item $\sum\limits_{t=1}^{T}a_{i,t} \leq 1$: each arm needs to be pulled at most once during the lookahead period. This is needed to make defining $w_{i,t}$ simple---otherwise $w_{i,t}$ depends on the time of the last pull.
    \item $a_{i,t} =  \left\{ 
                      \begin{array}{ c l }
                        1 \textrm{ or } 0 & \quad \textrm{if } \textrm{t\ in action window} \\
                        0                 & \quad \textrm{otherwise}
                      \end{array}
                    \right.$ This constraint forces $a$ out of the action window to be 0, which satisfies one of our problem setting: arms can only be pulled during their action windows.
    \item Additional desired frequency constraints, e.g., each arm must be pulled at least once during certain timesteps.
\end{enumerate}


\begin{prop}
Maximizing the sum of Whittle indices without additional frequency constraints OR with the constraint that each arm must be pulled exactly once during the lookahead window can be reduced to a weighted $b$-matching~\cite{doi:b_matching}.
\end{prop}
This is important because the number of arms in our motivating example can be very large, resulting in a large number of integer variables in the naive integer program (IP). See appendix for proof details.



In practice, it is convenient to solve this lookahead problem as an IP. We can do so with $NL$ binary variables $a_{i,t}$ (where $L$ is the length of the action window) and $T + N$ constraints. Because the polynomial tractability of weighted $b$-matching arises from total unimodularity~\cite{b-matching}, the IP can be solved very quickly via its LP relaxation. However, polynomial tractability is lost when sufficiently complex minimum and maximum number of pulls are added as additional constraints as the problem becomes equivalent to weighted bipartite $b$-matching~\cite{wbm}.

The IP can be extended to more complex cases, e.g., where there are multiple pulls for each arm in the lookahead window. To modify the Whittle index for a time $t'$ based on whether an arm was pulled at $t$ or not, we can add constraints of the form:
\begin{align} \label{eq:multipull} 
    w_{i,t'} \leq M(1 - a_{i,t}) + a_{i,t} w'_{i,t'}
\end{align}
where $w'_{i,t'}$ is the Whittle index at $t'$ for arm $i$ if it was pulled at time $t$. Note that Whittle indices will always decrease when a pull happens under our assumptions that an action improves the state of an arm. Thus, we can add $LN$ additional constraints to allow for an additional pull during the lookahead period.

\section{Action Window Optimization}
 \label{sec:window_determine}
In the problem of city and public service scheduling, including food establishment inspections, it is the authorized agency's duty to assign inspection windows. In the practice of food inspections, the establishments are aware of the window but not the exact inspection time. Under such circumstances, we will show that the techniques introduced in this paper can be leveraged to optimize window assignments to further increase rewards. In this setting, we still need to satisfy the desired service constraints (i.e., minimum and maximum number of pulls, no information provided about inspection time is provided beyond the window), but have the flexibility to place the windows as we choose. 

To accomplish this, we assign a ``virtual'' window to each establishment, consisting of the entire period during which the inspection constraints must apply. For example, if the desired outcome is exactly one inspection over the next year, we assign the virtual window of the entire year to each establishment. Then, we simulate the operation of the RMAB over the virtual window, using the techniques of Sec.~\ref{sec:satisfying} to ensure that the required constraints are satisfied and record when inspections occur, producing the \emph{virtual inspection sequence}. We then will take this sequence and use it to assign windows such that each virtual inspection takes place during the assigned window, and we will do so carefully to anonymize when the actual inspection will take place. Because we assume that inspections never fail to transition an establishment to the adherent state, there is no loss of expected reward incurred by executing the virtual inspection sequence determined during window planning.

How do we assign windows according to the virtual inspection sequence? Let $a_{i,t}$ be the encoding of the virtual inspection sequence, where $a_{i,t}=1$ if arm $i$ is pulled at time $t$. A naive way is to assign arm $i$ with window $[t, t + W - 1] $ if $a_{i,t} = 1$. However, such assignments are easily predictable and establishments would be able to prepare effectively. We need to design a way in which establishments get inspected with a probability of $1 / W$ on each day in the window. 

We can do so with a linear program (LP). We define variables $f_{t,t'}$, which indicate the proportion of arms with virtual inspection time $t$ that are put in inspection window $[t', t' + W - 1]$. We denote the number of arms with assigned inspection $t$ that are put in the window starting at $t'$ as $g_{t,t'}$.  Thus, $g_{t,t'} = \sum\limits_{i=1}^{N} a_{i,t}f_{t,t'}$.
For the objective, we use
\begin{align}
\min_{f} \sum\limits_{t=1}^{T}\sum\limits_{t'=t-W+1}^{t}\sum\limits_{t''=t-W+1}^{t}| g_{t,t'} -  g_{t,t''}|
\end{align}
to satisfy the anonymity condition---that the action window provides no additional information. We can use the standard constraint trick to remove the absolute value in the objective by introducing an auxiliary variable that is constrained to be larger than the objective and the negation of the objective. We add additional constraints to ensure that the window assignments achieve our goals:
\begin{enumerate}
    \item The virtual action assignment must occur during the window: $f_{t,t'} \neq 0 $ if and only if $0 \leq t - t' \leq W - 1$. 
    \item The window assignment probabilities sum to 1: $\sum_{t'} f_{t,t'} = 1$ for all $t$
\end{enumerate}
Using the $f_{t,t'}$ output of the LP, we sample windows using any categorical sampling procedure. At this stage, the virtual inspection can be discarded and individual inspections planned as if the windows were given (i.e., not even the system operator knows when the virtual inspections were planned to occur). In practice, unplanned inspections (i.e., due to complaints) will happen constantly, meaning replanning is necessary to satisfy constraints.

Optimizing the window positions has a larger effect than optimizing inspections within fixed windows in our experiments, which makes sense given the additional flexibility that window optimization affords. However, window optimization builds directly on our approach for optimizing inspections within fixed windows.

\section{Experimental Study}
In this section, we ask three questions. First, what is the impact on adherence of leveraging the methods of this paper to optimize inspection policies? Second, what is the cost of the inspection service constraints in terms of adherence? Third, how do different methods degrade when unplanned complaint inspections are occuring?
We describe the compared policies in Sec.~\ref{sec:policies} and study the impact of different planning policies on reward and computation time, both in synthetic (Sec.~\ref{sec:synthetic}) and real data from CDPH (Sec.~\ref{sec:real}) domains. 

\subsection{Planning Policies}\label{sec:policies}
We study different constraints situations from three dimensions: whether the window could be re-scheduled and optimized or random, whether the schedule is optimized (apply the methods in Sec~\ref{sec:satisfying} and Sec~\ref{sec:window_determine}), or naive IP, and the frequency constraints (at most once, exactly once, between once and twice per period). To the best of our knowledge, the constraint structures in this paper have not been studied before. Thus, there exist no prior methods for optimizing reward while satisfying the constraints. We thus compare to naive IP methods that guarantee constraint satisfaction, but are oblivious to reward.
\begin{itemize}
    \item \textbf{(Rdm, IP, =1)}: The action windows are distributed at random and each arm is pulled exactly once per year using an IP scheduler, satisfying all constraints. We think of this result as representing the status quo.
    \item \textbf{(Opt, IP, =1)}: We perform action window optimization (Sec~\ref{sec:window_determine}), but then apply the IP scheduler of the above policy.
    \item \textbf{(Rdm, Opt, =1)}: The action windows are distributed at random, each arm needs to be pulled exactly once per period, and we apply the method of Sec~\ref{sec:frequency} to optimize reward.
    \item \textbf{(Opt, Opt, =1)}: The action windows are optimized and each arm needs to be pulled exactly once per period. This policy is implemented by optimizing windows (Sec~\ref{sec:window_determine}) first and then applying the method of Sec~\ref{sec:frequency}.
    \item \textbf{(Rdm, Opt, $\boldsymbol{\leq}$ 1)}: The action windows are distributed at random, each arm needs to be pulled \emph{at most once} per period, and the method of Sec~\ref{sec:window} is used to optimize rewards.
    \item \textbf{(Opt, Opt, $\boldsymbol{\leq}$ 1)}: The action windows are optimized and each arm needs to be pulled at most once per period. We optimize windows, and then we model the RMAB as Sec~\ref{sec:window}.
    \item \textbf{(Opt, Opt, [1,2], Budget\%)}: We want to study the benefits of additional inspection budgets. Thus 10\%, 12\%, and 15\% budget experiments are conducted under the constraint that each establishment is inspected between one and two times. This policy is implemented by optimizing windows (Sec~\ref{sec:window_determine}) first and then using the method of Sec~\ref{sec:frequency} with Equation~\ref{eq:multipull} for the purpose of multiple pulls in a period.
    \item \textbf{(Opt, IP, [1,2], Budget\%)}: Baseline for the previous setting, where we apply the window optimization using the method in Sec~\ref{sec:window} and then apply an IP scheduler with frequency constraints (at least once, at most twice).
\end{itemize}

The same method is used to compute Whittle indices for all policies that require them. See Appendix for detailed computation costs.


\paragraph{Measuring Reward}
We measure the reward value in expectation, summing the \emph{probability} of adherence across time periods for each establishment. This makes individual runs deterministic---the only stochasticity is in the experimental setup is the parameters of the establishments in the synthetic domain.

\subsection{Synthetic Domain}\label{sec:synthetic}
We begin with experiments using synthetic instances.

\subsubsection{Data Preparation and Setup}
In the synthetic domain, we generate $P^{(0)}_i[0,0]$ by sampling from $\textrm{Beta}(\alpha=5, \beta=1)$ and $P^{(0)}_i[1,0]$ by sampling from $\textrm{Beta}(\alpha=1, \beta=5)$. 
Each simulation is run for 60 timesteps (``months''). Each arm has two action windows of two months each, with one pull allowed in each window.
We set the number of arms to 1000 and the budget per round to 9\% of the number of arms (we need a budget of 8.33\% of all arms to satisfy all arms' constraints).

\subsection{Food Establishment Inspection Domain}\label{sec:real}
Using inspection data from the Chicago Data Portal~\cite{foodinspectiondata}, we implement a realistic RMAB setting.

\subsubsection{Data and Setup}
Since 2010, CDPH has published every food establishment inspection result on the Chicago Data Portal~\cite{foodinspectiondata}. The Food Inspection Dataset is a tabular dataset with 17 attributes for each establishment including license number, address, inspection results, etc. A neural network with 2 MLPs is trained to predict the transition probabilities. More details of data and model structures are shown in the appendix.




\paragraph{Loss of Model Training}
The loss is to minimize the negative log likelihood:
\begin{align}
    \min_{\theta} \sum_y {-\log(p(s';s \theta^{T}))}
\end{align}
where $s$ is the last state and $s'$ is the next state, $T$ is the time since last inspection.
The Adam optimizer is applied and the learning rate is 0.0001, and the model is trained for 10000 epochs.

\paragraph{MLP Training Result}
We validate the MLP by computing the AUC of its predictions. To do this, we view each interval between predictions as a data point with the label of whether the next inspection found adherence or non-adherence. If no previous records for the establishment, we use average values to fill missing columns. We then take the parameter predicted by the MLP and use it to compute the probability of adherence and compute the resulting AUC. Then we trained the model on different train-test splits:
\begin{itemize}
    \item Split all inspection data randomly, holding out 20\%: AUC of \textbf{0.74}.
    \item Split inspection data by establishments (holding out 20\% of establishments for testing): AUC of \textbf{0.75}.
\end{itemize}
The fit to real-world data is excellent compared with the XGBoost model with an AUC of 0.67 in the~\cite{foodforecast}.

In the test experiments, we have 1801 arms, 60 timesteps, and a 9\% budget. When inspection windows are set randomly, we use the same windows of two consecutive steps per year as in the synthetic data.

\subsection{Results}

\begin{figure}[t]
\centering
\includegraphics[width=1\columnwidth]{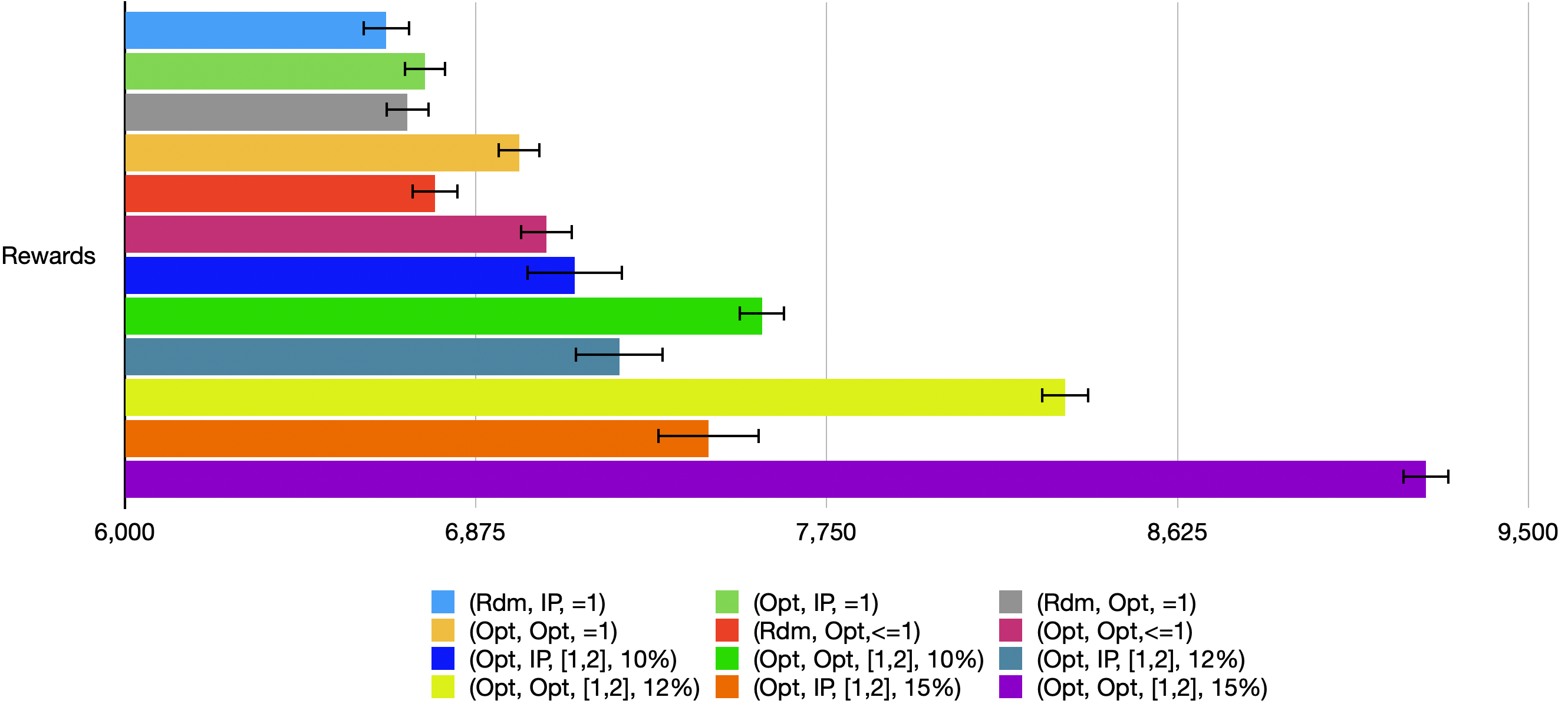} 
\caption{Results from the synthetic domain (with standard errors).}
\label{fig:syn_rel}
\end{figure}

\begin{figure}[t]
\centering
\includegraphics[width=1\columnwidth]{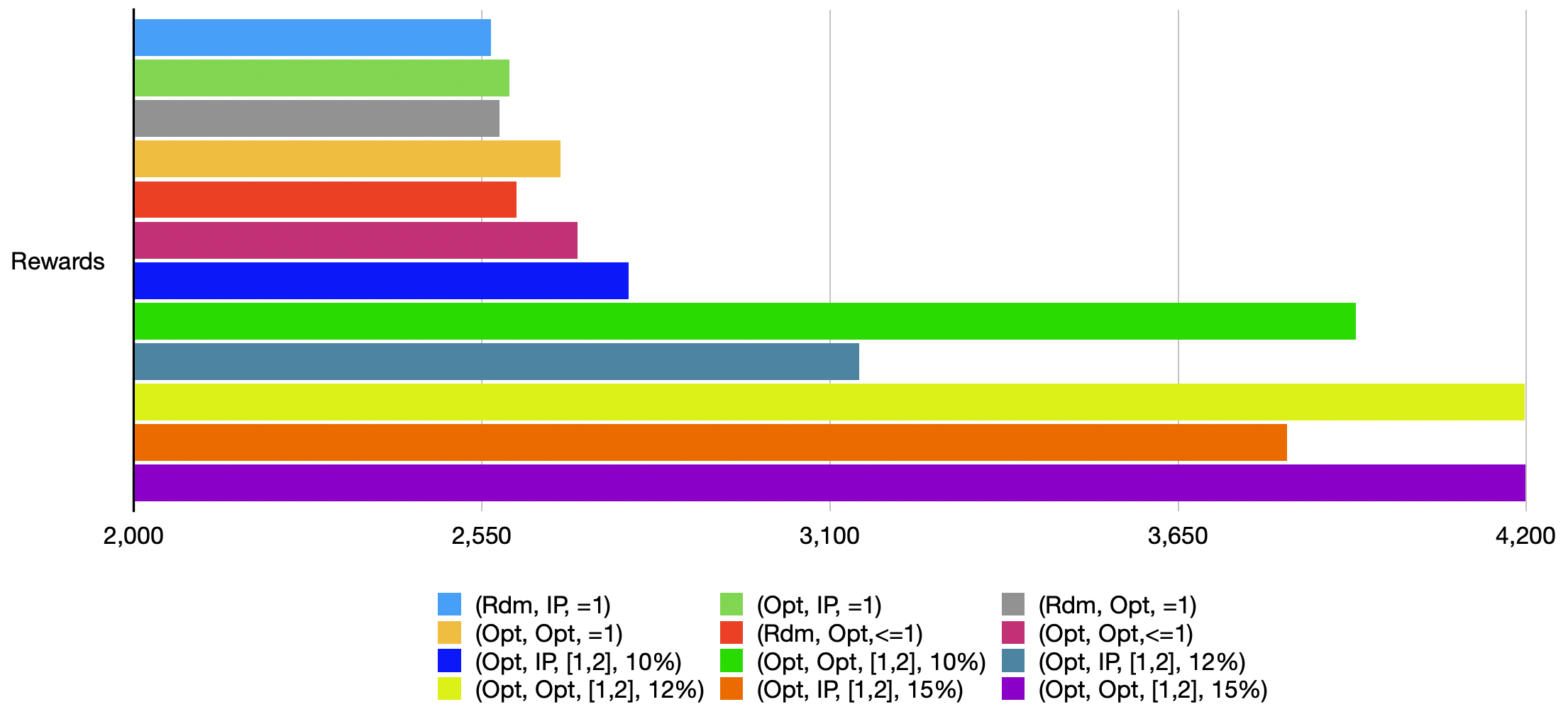} 
\caption{CDPH domain results. There are no error bars because the CDPH data defines a single model.}
\label{fig:real_rel}
\end{figure}

\begin{figure}[t]
\centering
\includegraphics[width=1\columnwidth]{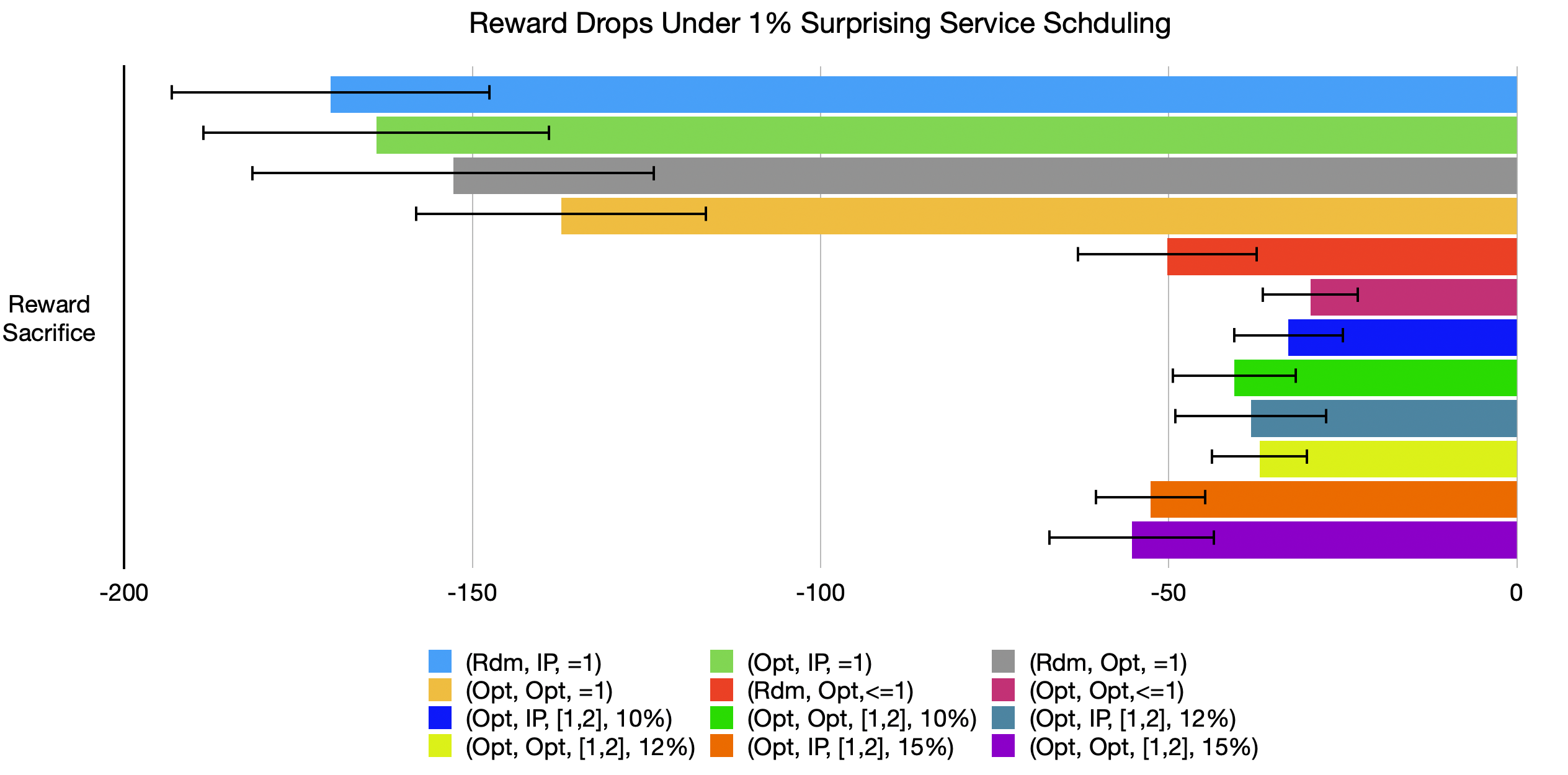} 
\caption{The reward drop caused by introducing surprise inspections at a 1\% rate compared in the synthetic domain relative to the rewards in Figure~\ref{fig:syn_rel}.}
\label{fig:sur}
\end{figure}

\begin{table}[]
\centering
\begin{tabular}{lll}
\multicolumn{3}{c}{Average of Difference}                                  \\ 
\multicolumn{1}{l|}{} & (Opt, Opt, =1) & (Opt, IP, =1)                         \\ \hline
\multicolumn{1}{l|}{N(0, 0.05)} & -83.69 ± 20.14 & -101.14± 35.55                         \\
\multicolumn{1}{l|}{N(0, 0.1)}  & -86.43 ± 69.99  & 21.11 ± 100.62                   \\
\multicolumn{1}{l|}{N(0, 0.15)} & -101.23 ± 82.51 & -130.23 ± 85.19 \\
\multicolumn{1}{l|}{N(0, 0.20)} & -151.88 ± 73.79  & -170.27 ± 84.52                   
\end{tabular}
\caption{Experiment results for robustness study.}
\label{tbl:noise_rel}
\end{table}

The total reward accrued for each policy is presented in Figure~\ref{fig:syn_rel} as reward improvements relative to the reward achieved by pulling no arms at all. The benefits of applying the methods of this paper are clearly seen---optimizing schedules within a fixed window (Rdm, Opt, =1) receives a higher reward than the baseline (Rdm, IP, =1), and optimizing the windows and schedules (Opt, Opt, =1) is better still. In addition, there is substantial synergy between window optimization and schedule optimization---when windows are optimized, but the schedule is not (Opt, IP, =1), the reward is much less than when both are optimized together (Opt, Opt, =1). The most impactful scenario for our methods is when the algorithm can select some establishments to inspect twice yearly, while still meeting the constraints on all other establishments, with an improvement of up to 24\% in the synthetic domain and 33\% in the CDPH domain.


The CDPH domain results are shown in Figure~\ref{fig:real_rel}. The ordering of the results is the same, with less improvement overall relative to no inspections. Most food establishments tend to stay in the adherent state, and thus, the benefit of inspections is not as significant as the synthetic domain. We also observe budget saturation in the additional budget scenarios, with 12\% and 15\% budget yielding the same reward under optimization.

In practice, there will be errors in the estimates of the establishments' transition probabilities. We perform a sensitivity analysis on this error via adding Gaussian noises on the parameters. We compare our method (Opt, Opt, =1) with the baseline (Opt, IP, =1). The result shows the difference of rewards before and after adding noise in Table~\ref{tbl:noise_rel}. We find that zero-mean Gaussian noise in the exactly one inspection setting in the synthetic domain reduces reward by around 100 points for both (Opt, Opt, =1) and (Opt, IP, =1), and we find that the reward is reduced by the same or less in (Opt, Opt, =1) vs.\ (Opt, IP, =1).

CDPH is also responsible for unscheduled inspections, e.g., in response to complaints. We conducted simulation experiment assuming each food establishment has 1\% probability of an unscheduled inspections and evaluated the impact to the rewards to the synthetic setting of Fig.~\ref{fig:syn_rel} and show the results in Fig.~\ref{fig:sur}. We saw both our window and scheduling optimization shows robustness compared with IP and Random policies. With \textbf{=1} constraint, a regular IP would suffer around 2.6\% reward losses while ours only sacrifices 1.9\%. If \textbf{=1} is removed, reward loss would drop less than 0.5\% and the gap between our method and IP is narrowing, showing that strict frequency constraints have an essential impact to rewards.

Our results also provide insight into the cost of the service constraint that each establishment should be inspected exactly once per year, rather than at most once. We find the cost of this constraint to be moderate, roughly of the same magnitude as going from fixed to optimized windows or from a random to optimized inspection schedule under fixed windows. However, relaxing this constraint has a substantial cost---that not every establishment is inspected each year. This weakens the guarantee to the consumer that every part of the distribution chain is inspected periodically. In contrast, window and schedule optimization do not require any weakening of this guarantee.

\section{Conclusions}
We propose an RMAB-based approach for real-world urban service scheduling. Experiments on synthetic data and Chicago Department of Public Health inspection data show that explicitly modeling constraints and optimizing action windows are crucial for RMAB effectiveness. This work paves the way for applying RMABs to other constrained public service and infrastructure maintenance problems.

\bibliography{aaai2026}

\appendix

\section{Implementation Details}
Additional details and implementation code are available at:
\url{https://github.com/Peter-beeler/TimeConstrainedBandit}.
\section{Inspection Window and MDP Details}
We argue that an unanticipated inspection assignment(like ours) is essential in such a public health check. A study by~\citet{surprise_ins} exploits variation in inspection predictability among co-located food-service establishments in LA County to assess how anticipation affects regulatory compliance. Establishments receiving unanticipated sole inspections exhibit significantly worse compliance, with 7.75\% more violations and 16.3\% more major critical violations than those inspected concurrently with neighboring establishments. This studies show establishments are preparing for inspections meticulously which does not provide insight into critical violations outside the temporal scope of inspections.

To enforce that there are $\eta$ timesteps of no action (sleep) after each action, as in ~\citet{field_study}, requires a factor of $\eta$ more states. We add a counter to the state to record the number of timesteps until the next pull is allowed. When the counter is positive, the effect of an action is the same as the effect of no action.

This means the Whittle heuristic will never select arms outside the action window (or during mandatory sleep) as long as there is some arm with positive action effects. This is because the Whittle index of arms that are not eligible to be pulled is zero as arms outside the action window (or during mandatory sleep) have no advantage for the active action over the passive action. If they are selected anyway, the agent can discard these actions to no ill effect.

\section{Algorithm Computational Costs}

\begin{table}[t]
\centering
\begin{tabular}{ll}
    Policy       & Time (sec)       \\ \hline
(Rdm, IP, =1)    & $88.49\pm4.05$ \\
(Opt, IP, =1)   & $1070.21\pm44.54$ \\
(Rdm, Opt, =1)    & $1106.35\pm 52.31$ \\
(Opt, Opt, =1)    & $1072.24\pm45.73$ \\
(Rdm, Opt, $\leq$1)   &  $1028.19\pm51.97$\\
(Opt, Opt, $\leq$1)   & $982.34\pm44.54$ \\

\end{tabular}
\caption{The average running time in seconds for a 1000-arm RMAB over 12 steps. }
\label{table: time_compare}
\end{table}

We simultaneously compute Whittle indices for all states of each arm using binary search over subsidies with the tolerance $10^{-6}$. All experiments are run on a single core of AMD EPYC 7643 (Milan) processors (2.3 GHz), on Ohio Supercomputer Center~\cite{Ohio_Supercomputer_Center1987-dl}.
For an RMAB with 1000 arms, computing Whittle Indices takes around 1000s and the baseline scheduling IP takes around 100s for one period. Details of the running time of policies can be seen in Table~\ref{table: time_compare}. Optimizing policies consumes around an order of magnitude more computational time than the baseline, but we can reuse the Whittle indices for window optimization for scheduling, and as a result, optimizing windows and schedules together requires negligibly more computation time than performing either optimization individually. RAM consumption is low for all policies, less than 500MB.

\section{Transition Inferring Model and Data Preparation}
\paragraph{Data Preparation}
The inputs to the model are a series of features of establishments. We use the same features as previous work on predicting food establishment risks ~\cite{foodforecast}, which combines data of business licenses, food inspections, crime, garbage complaints, sanitation complaints, weather and sanitation information.
The inspection results are shown in the ``Violations'' column: 0 means no violations and pass, 1 means violations appear and 2 means pass with conditions. In the experiment, both 0 and 2 are merged into a single good state and 1 is the bad state. We group all inspection data by establishments and transfer them to a transition trajectory data which contains original features, start/end states and time spans.

\begin{figure}[t]
\centering
\includegraphics[width=1\columnwidth]{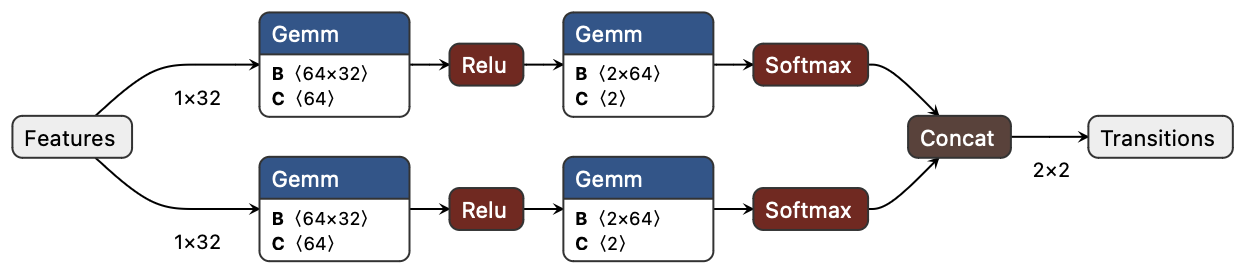} 
\caption{Structure of neural networks of learning transitions of food establishments.}
\label{fig:model_trans}
\end{figure}

\paragraph{Inferring Transitions}
To create a realistic instance, we must infer the transition probabilities from the inspection trajectories for each arm. 
Because of the small number of inspections per establishment, it is impractical to estimate a transition model for each food establishment. Thus, we used all data to train one single model to learn the transitions for all establishments. A neural network with 2 MLPs is trained to predict the transition matrix $P^{(0)}_i$ to maximize the log-likelihood of the data given the transitions. The detailed architecture of the neural network can been seen in Figure~\ref{fig:model_trans}.
\section{Weighted $b$-Matching}
The lookahead planning algorithm we develop will be reducible to variants of the weighted $b$-matching problem~\cite{b-matching}. A weighted $b$-matching instance is described by an undirected graph $G=(V,E)$, an edge weight vector $w: E \rightarrow \mathbb{R}$, and a non-negative $b$ vector $b: V \rightarrow \mathbb{N}_+$. The objective in a maximum weight $b$-matching is to find a set of edges $x$ with maximum weight, subject to the constraint that only $b(v)$ edges that are adjacent to node $v$ can be selected. Formally, 
\begin{align}
    \max_x w^T x, \hspace{10pt} \textrm{s.t. } \sum_{u} x_{u,v} \leq b(v), \forall v \in V
\end{align}
Weighted $b$-matchings can be solved in polynomial time, e.g., in $O( |V|^2 \max_v b(v))$~\cite{pulleyblank_matching}.

A more challenging weighted $b$-matching variant is weighted bipartite $b$-matching~\cite{wbm}. In this variant, graph nodes are partitioned into a right set $U$ and left set $V$, and there are no edges within each partition. Nodes in the left (resp., right) set have maximum matching cardinality $L^+$ (resp., $R^+$) and minimum cardinality $L^-$ (resp., $R^-$). Under these constraints, finding a maximum weight $b$-matching is NP-hard.

\subsubsection{Proof For Proposition 1}
\begin{proof}
The proof converts each timestep and each arm to nodes in the matching graph with different $b$-values. 
We formulate the weighted $b$-matching instance as follows. For each arm $i \in [N]$, create a node $i$. For each timestep $t$ in the lookahead period, create a node $t$. For each arm-timestep pair $(i, t)$ where an action can occur (i.e., no timing constraints are violated), create an edge of weight $w_{i,t}$ between $i$ and $t$. Set the $b(t)=k$ for all $t$ and $b(i)=1$ for all nodes $i$. We claim that the maximum weight $b$-matching can be converted to an optimal lookahead schedule by taking each arm-timestep $(i,t)$ pair that is included in the maximum weight $b$-matching and pulling the arm $i$ at timestep $t$. Constraints 1 and 2 are satisfied by definition of weighted $b$-matching. Constraint 3 is satisfied because edge $(i,t)$ exists only if $t$ is in $i$'s action window. Thus, the optimal solution to the weighted $b$-matching must be the optimal solution to the lookahead problem.

To account for the additional frequency constraint that each arm must receive at least one pull in the lookahead window, if possible, a large constant can be added to all Whittle indices. The constant will cause each arm to be pulled once, if possible, because it is much larger than the increase in objective value that can be achieved by shifting the pull time for any individual arm.
\end{proof}

\section{Additional Experiment Results}
\subsection{Restore Failure}
In our baseline model, we assume perfect restoration following inspection, as the regulatory system prohibits establishments from reopening after critical violations are identified. To assess the robustness of our framework, we examine a scenario in which restoration may fail stochastically. Specifically, we modify the active transition matrix $P_i^{(1)}(0,0)$ from 1 to 0.95, representing a 5\% probability that an inspection fails to restore an establishment to a compliant state. All other components of the framework, including the Whittle index computation and the optimization window, remain unchanged. We evaluate the policy
(Opt, Opt, =1) under both restoration rates. The results show that perfect restoration (100\%) yields an reward of 41898.92, while the 95\% restoration rate produces 41894.01. This difference of approximately 0.01\% is negligible and demonstrates that our algorithm remains robust to small levels of restoration failure.
\subsection{Hyper-paprameter Finetuning}
For the $\alpha$ and $\beta$ choice in Sythetic Domain, to ensure our results are not driven by these “magic” choices, we performed a systematic parameter-finding and sensitivity analysis. We fit Beta($\alpha, \beta$) priors to the empirical transition statistics from the CDPH data and used those fitted values as our canonical synthetic settings($P_i^{(1)}(0,0)$ ) using Scipy and it returns $\alpha=5.08$ and $\beta=0.52$. For simplicity and symmetry, we choose Beta(5,1) for $P_i^{(1)}(0,0)$ and Beta(1,5) for $P_i^{(1)}(1,0)$. We tested robustness by adding zero-mean Gaussian noise to transition parameters in Table~\ref{tbl:noise_rel}. Meanwhile, we also tested the sensitivity to $\alpha$ and $\beta$ by also adding Gaussian noise with zero-mean and 0.1 variance on a smalle-scale data generation and experiments. Results shows the mean of \textbf{4041.04}  with a standard deviation of \textbf{254.78}, which indicates relatively stable performance.

\section{Fairness Study}
Our work proposes algorithms to optimize inspection allocation and prioritization using historical public food-establishment inspection data. While the primary goal is to improve efficiency and public safety, automated decision systems of this type can also produce equity-related effects, both positive and negative.

\paragraph{Potential benefits}
By formalizing inspection scheduling as a transparent, data-driven optimization problem, our framework can help reduce arbitrary or ad-hoc decision making. When paired with publicly available data and interpretable models, this may enhance procedural fairness by making the criteria for public service scheduling more consistent. Moreover, constraints represented by optimization problems ensure the a fair frequency of disturbance for businesses. 

\paragraph{Potential risks}
However, the approach also inherits the biases present in historical inspection data. Past inspection patterns in cities can reflect structural inequities—for example, in~\citet{foodforecast}'s study, the inspector has a great factor weight in XGBoost result, this may also cause some biases our transition-predicting model though consequences are not supposed to be catastrophic due to our constraints.

\paragraph{Mitigation strategies}
To address this, future work should incorporate fairness constraints or equity-aware regularization into the optimization objective, ensuring that inspection frequency or expected penalties are balanced across demographic or geographic subgroups. Periodic auditing and transparency reports could monitor the distribution of inspections and outcomes by neighborhood, establishment size, or ownership demographics. Furthermore, participatory feedback from local stakeholders can guide the definition of fairness criteria that align with community values.

In sum, while our method can improve the effectiveness of public health inspection systems, its deployment must be accompanied by deliberate equity safeguards—including bias audits, and multifaceted evaluations during the operation.

\section{Demographic Fairness IP}
To study equity and resource-allocation balance, we extend the planning by adding fairness constraints that restrict the allocation of service to each demographic or geographic subgroup (e.g., each neigborhood).

Let $\mathcal{I}$ be the set of agents (establishments), 
$\mathcal{G}\subseteq\mathcal{I}$ denote a demographic or geographic group (e.g., a ward),
and $T$ the planning horizon.  
For each $i\in\mathcal{I}$ and $t\in\{1,\dots,T\}$. Let $B_t$ be the inspection budget at time $t$.

The fairness-constrained look-ahead integer program is

$$
\begin{aligned}
\max_{a}\;&
   \sum_{t=1}^{T}\sum_{i\in\mathcal{I}} w_{i,t}\, a_{i,t} \\[4pt]
\text{s.t.}\;&
   \sum_{i\in\mathcal{I}} a_{i,t} \le B_t,
      &&\forall t=1,\dots,T, \\[4pt]
&
   \sum_{i\in\mathcal{G}}\sum_{t=1}^{T} a_{i,t}
      \ge
      \lambda
      \sum_{i\in\mathcal{I}}\sum_{t=1}^{T} a_{i,t},
      &&\lambda\in[0,1], \\[4pt]
&
   a_{i,t}\in\{0,1\},
      &&\forall i\in\mathcal{I},\, t=1,\dots,T.
\end{aligned}
$$

Constraint (2) enforces that group $\mathcal{G}$ receives at least a $\lambda$-fraction of total inspection attention over the horizon $1\!:\!T$, 
i.e., a demographic-parity or minimum-coverage requirement. 
For the different fraction requirements in a group, $\mathcal{G}_k,\;k=1,\dots,K$ with individual targets $\lambda_k$, constraint (2) would be:

$$
\sum_{i\in\mathcal{G}_k}\sum_{t=1}^{T} a_{i,t}
   \ge
   \lambda_k
   \sum_{i\in\mathcal{I}}\sum_{t=1}^{T} a_{i,t},
   \qquad \forall k=1,\dots,K.
$$
Noted our IP look-ahead~\ref{sec:frequency} is capable of encoding many such fairness constraints over the look ahead window, illustrating the extensibility of our model.

\end{document}